\documentclass[10pt]{article}

\usepackage{times}
\usepackage{geometry}
\usepackage{graphicx}
\usepackage{subcaption}
\usepackage{booktabs}
\usepackage{multirow}
\usepackage{url}
\usepackage{hyperref}
\usepackage{float}
\usepackage{natbib}
\usepackage{microtype}
\usepackage{tikz}
\usepackage{tabularx}
\usepackage{array}

\usepackage{amsmath,amsfonts,bm}

\def\eqref#1{equation~\ref{#1}}

\def\1{\bm{1}}

\DeclareMathAlphabet{\mathsfit}{\encodingdefault}{\sfdefault}{m}{sl}
\SetMathAlphabet{\mathsfit}{bold}{\encodingdefault}{\sfdefault}{bx}{n}

\usetikzlibrary{shapes,arrows.meta,positioning,calc,fit,backgrounds}

\title{SciFig2Data: A Workflow for Recovering Structured Data from Scientific Figures}

\author{
Xinhe Wu \\
iSoftStone AI Research Lab (ISSAIR)\\
iSoftStone\\
Beijing, China \\
\texttt{xhwude@isoftstone.com}
\and
Yicheng Ye \\
iSoftStone AI Research Lab (ISSAIR)\\
iSoftStone\\
Guangzhou, China \\
\texttt{ycyeh@isoftstone.com}
\and
Yadong Jin\thanks{Corresponding author.}\\
iSoftStone AI Research Lab (ISSAIR)\\
iSoftStone\\
Beijing, China \\
\texttt{ydjink@isoftstone.com}
}

\begin{document}

\maketitle

\begin{abstract}
Scientific measurements reported in scholarly articles are often embedded in compound figures rather than available as machine-readable data. Recovering these measurements at literature scale requires an integrated workflow for figure localization, panel decomposition, chart identification, numerical reconstruction, and experimental context association. We present \textit{SciFig2Data}, an automated, high-throughput workflow for harvesting structured experimental data from scientific PDF repositories. The system integrates multi-scale document element detection, adaptive compound figure decomposition, two-stage OCR-based semantic filtering, Swin-based chart classification, and instruction-guided multimodal large language model (MLLM) reconstruction. It further incorporates directory-level batch execution, caching, and structured logging to support large-scale processing. We demonstrate the workflow on two biomedical literature corpora comprising 605 toxicology and pharmacology articles and 551 lipid nanoparticle (LNP) articles. The pipeline successfully reconstructed 468 and 242 tabular outputs from the respective corpora, illustrating its practical applicability to literature-scale scientific data harvesting. By associating reconstructed measurements with caption-derived experimental metadata, SciFig2Data provides a workflow for transforming graphical evidence in scholarly publications into structured, context-enriched data records for downstream AI4Science applications.
\end{abstract}


\section{Introduction}

The rapid development of Artificial Intelligence for Science (AI4Science) has increased the demand for large-scale, structured scientific datasets. However, substantial experimental evidence remains embedded in graphical visualizations within scholarly publications, where numerical values are not directly accessible in machine-readable formats~\cite{wang2023scientific,davila2021chart}. Recovering these data is essential for building reusable scientific databases and supporting data-driven discovery.

Traditional figure digitization relies on manual or semi-automated tools, requiring researchers to identify charts, interpret axes and legends, and convert pixel coordinates into numerical values~\cite{pick2019metadigitise,luo2021chartocr}. Recent multimodal large language models (MLLMs) offer opportunities to automate this process through joint visual and textual understanding~\cite{liu2023deplot,islam2024chartcomprehension,liu2026start}.

Reliable numerical reconstruction remains challenging for visually dense scientific figures containing multiple panels, experimental conditions, and data series~\cite{davila2021chart}. 
Directly processing such composite figures requires
simultaneous panel identification, semantic interpretation,
and numerical reconstruction, increasing the risk of
incomplete extraction and numerical inaccuracies.
The previous work, \textit{ChartDensity-Bench}~\cite{wu2026chartdensity}, demonstrated that increasing visual density generally degrades MLLM reconstruction performance across multiple models, affecting numerical fidelity and extraction completeness. These findings motivate structured preprocessing to reduce visual interference before numerical reconstruction.

Beyond individual chart reconstruction, literature-scale data harvesting requires an integrated workflow that handles document element detection across heterogeneous PDF layouts~\cite{pfitzmann2022doclaynet}, compound figure decomposition, semantic chart identification, and numerical recovery. Existing studies have advanced chart understanding and reconstruction~\cite{davila2021chart,luo2021chartocr,yuan2026chartrecover}, but reusable scientific datasets also require contextual information such as experimental variables, units, and sample information~\cite{pick2019metadigitise}.

To address these challenges, we develop an end-to-end workflow for harvesting numerical experimental data from scholarly PDFs. 
Building on the observations from
\textit{ChartDensity-Bench}, this work shifts the focus
from model-level evaluation to an engineering-oriented
document-to-data workflow, integrating structured
preprocessing, instruction-guided numerical reconstruction,
and caption-based metadata enrichment.

The main contributions of this work are summarized as follows:
\begin{itemize}
    \item \textbf{An integrated literature-to-data workflow.}
    We develop a modular pipeline that transforms heterogeneous scholarly PDFs into structured experimental data through document parsing, visual analysis, numerical reconstruction, and contextual information recovery.
    \item \textbf{Structured preprocessing and instruction-guided reconstruction.}
    We integrate multi-scale document element detection, compound figure decomposition, spatial reading-order restoration, semantic filtering, and chart-specific MLLM prompting to organize and recover quantitative information.
    \item \textbf{An engineering demonstration.}
    We illustrate the end-to-end workflow through a scientific literature case study, demonstrating the transformation of compound figures into structured, context-enriched experimental records.
\end{itemize}

\section{System Overview}
\label{sec:system_overview}

The proposed framework transforms heterogeneous scholarly PDFs into structured experimental data through six stages: document element detection, pre-decomposition semantic OCR filtering, compound figure decomposition, chart classification and panel-level OCR filtering, instruction-guided numerical reconstruction, and caption-based metadata integration. Table~\ref{tab:system_architecture} summarizes the inputs, core methods, and outputs of each stage.

\begin{table*}[t]
\centering
\caption{Overview of the proposed literature-to-data workflow and its major processing modules.}
\label{tab:system_architecture}
\small
\renewcommand{\arraystretch}{1.3}
\begin{tabularx}{\textwidth}{
  >{\raggedright\arraybackslash}p{0.18\textwidth}
  >{\raggedright\arraybackslash}p{0.20\textwidth}
  >{\raggedright\arraybackslash}X
  >{\raggedright\arraybackslash}p{0.20\textwidth}
}
\toprule
\textbf{Module} & \textbf{Input} & \textbf{Core Method} & \textbf{Output} \\
\midrule
Document Layout Detection & Scholarly PDF pages & Multi-scale rendering and DocLayout-YOLO detection & Candidate figure and table regions \\

Pre-Decomposition Semantic Filtering & Cropped candidate figures and tables & Adaptive binarization, dual-orientation Tesseract OCR, and co-occurrence keyword gating & Domain-relevant compound figures and serialized table texts \\
Compound Figure Decomposition & Filtered compound figures & Subfigure separation, boundary expansion, and reading-order restoration & Individual subfigure panels \\
Semantic Filtering and Chart Classification & Decomposed subfigure panels & Panel-level OCR refinement and Swin-chart classification & Relevant quantitative chart panels \\
Instruction-guided Numerical Reconstruction & Classified chart panels and structured instructions & Chart-specific MLLM reconstruction & Structured numerical records \\
Caption-based Metadata Integration & Reconstructed numerical records and figure captions & Caption-based metadata association & Context-enriched experimental data \\
\bottomrule
\end{tabularx}
\end{table*}

\subsection{Document Layout Detection}
\label{subsec:doc_element_detection}

Each PDF page is rendered at two resolutions to identify
candidate graphical elements while preserving sufficient
visual information for subsequent cropping and analysis. The detection-scale images are analyzed by \texttt{DocLayout-YOLO}, a YOLOv10-based document layout detector~\cite{zhao2024doclayoutyolo}, to localize candidate figures and embedded tables.

The detected bounding boxes $(x_1,y_1,x_2,y_2)$ are subsequently projected into the cropping coordinate space. To preserve peripheral visual information, including axis titles, tick labels, and legends, each bounding box is expanded using a context padding parameter $\rho$:
\begin{align}
    x_1' &= \max(0,\lfloor x_1 s-\rho \rfloor), & y_1' &= \max(0,\lfloor y_1 s-\rho \rfloor), \\
    x_2' &= \min(W,\lceil x_2 s+\rho \rceil), & y_2' &= \min(H,\lceil y_2 s+\rho \rceil),
\end{align}
where $s$ is the scaling factor from detection to cropping coordinates, and $W$ and $H$ denote the dimensions of the cropping-scale image.

\subsection{Pre-Decomposition Semantic Filtering}
\label{subsec:pre_decomp_filtering}

Prior to compound figure decomposition, an initial semantic screening pass eliminates non-target graphics (e.g., microscopy images or schematic diagrams) to reduce downstream computational overhead. Each cropped candidate image undergoes a standardized preprocessing pipeline—comprising grayscale conversion, bilateral filtering, $2\times$ bicubic upscaling, and Gaussian adaptive thresholding—to enhance text contrast and suppress background noise. 

To capture both horizontal annotations and vertical axis labels, Tesseract OCR (\texttt{eng+chi\_sim}) is executed on both the binarized image $I_{\text{binary}}$ and its $90^\circ$ clockwise rotation:
\begin{equation}
    T_{\text{full}} = \text{OCR}(I_{\text{binary}}) \mathbin{\Vert} \text{OCR}(\text{rot}_{90}(I_{\text{binary}})).
\end{equation}
The recognized text is normalized (lowercased and stripped of punctuation and whitespace anomalies) and evaluated against a domain-specific lexicon $\mathcal{L} = \{\mathcal{K}_{\text{independent}}, \mathcal{K}_{\text{dependent}}\}$ representing prerequisite experimental variables (e.g., \textit{dose}, \textit{concentration}) and assay readouts (e.g., \textit{viability}, \textit{death}, \textit{ATP}). A candidate figure is retained for subfigure decomposition only when both keyword groups co-occur:
\begin{equation}
    \mathbb{I}_{\text{stage1}} = k_1 \land k_2,
\end{equation}
where $k_1$ and $k_2$ denote matches from $\mathcal{K}_{\text{independent}}$ and $\mathcal{K}_{\text{dependent}}$, respectively. Extracted table regions undergo the same OCR pipeline and are serialized directly into structured JSON records for downstream reference.

\subsection{Compound Figure Decomposition}
\label{subsec:compound_chart_decomposition}

Retained scientific figures often contain multiple panels (e.g., Panels A--F), whose simultaneous processing may introduce visual interference and complicate data-series association. We adopt the YOLOv2-based subfigure separation model and its pretrained weights released by Tsutsui and Crandall~\cite{tsutsui2017compound,redmon2017yolo9000}. The detected sub-panels are then processed by two additional post-processing steps developed in our workflow to improve panel preservation and spatial consistency.

First, adaptive boundary expansion is applied to preserve peripheral visual information that may be excluded by the initial segmentation. The structural complexity of a candidate figure is estimated using Canny edge density:
\begin{equation}
    D_{\text{edge}} = \frac{\sum_{x,y} E(x,y)}{W H},
\end{equation}
where $E(x,y)$ is the binary edge map and $W$ and $H$ are the image dimensions. When the edge density exceeds 1\%, panel boundaries are
expanded by up to 10\% along each dimension to better
preserve peripheral elements, including axis titles
and tick labels.

Second, spatial reading-order restoration is performed to recover the natural arrangement of the decomposed panels. Given two candidate panels $i$ and $j$, their vertical overlap ratio is defined as:
\begin{equation}
    R_{\text{overlap}}(i,j) = \frac{\operatorname{height}(B_i \cap B_j)}{\min(h_i,h_j)},
\end{equation}
where $B_i$ and $B_j$ are the panel bounding boxes, and $h_i$ and $h_j$ are their heights. Panels are assigned to the same row if their overlap exceeds $\tau_{\text{overlap}}$ or their vertical center distance satisfies:
\begin{equation}
    \left| y_{\text{center}}^{(i)} - y_{\text{center}}^{(j)} \right| < \delta \cdot \frac{h_i+h_j}{2}, \qquad \delta = 0.2.
\end{equation}
Row clusters are sorted vertically, and panels within each row are sorted by their left coordinates to restore top-to-bottom, left-to-right reading order. Figures yielding fewer than two panels are retained as single visual units.

\subsection{Semantic Filtering and Chart Classification}
\label{subsec:chart_classification_and_ocr}

Because compound figures filtered in Section~\ref{subsec:pre_decomp_filtering} may still contain auxiliary non-quantitative panels alongside target plots, we apply a secondary panel-level OCR pass followed by visual chart classification.

\paragraph{Panel-Level Semantic OCR Filtering.}
Following decomposition, each individual subfigure panel is re-evaluated via OCR using multilingual character sets, including mathematical and Greek symbols. In contrast to the strict figure-level co-occurrence criterion ($\mathbb{I}_{\text{stage1}} = k_1 \land k_2$), a relaxed retention rule is applied at the panel level:
\begin{equation}
    \mathbb{I}_{\text{stage2}} = k_1 \lor k_2.
\end{equation}
Whereas the strict initial pass eliminates irrelevant whole figures early in the pipeline, the relaxed panel-level criterion preserves relevant subfigures that carry only partial axis annotations while discarding unrelated auxiliary panels prior to numerical reconstruction.

\paragraph{Visual Paradigm Classification.}
Panels passing semantic filtering are classified using a fine-tuned Swin-chart model~\cite{dhote2024swinchart} based on the Swin Transformer architecture~\cite{liu2021swin}. Five paradigms are recognized: horizontal bar, vertical bar, line, scatter, and scatter-line. Panels with confidence $p \geq 0.7$ are forwarded to reconstruction, with the predicted paradigm determining the extraction protocol.

\subsection{Instruction-guided Numerical Reconstruction and Integration}
\label{subsec:instruction_guided_recovery}

To improve numerical reconstruction, we employ
chart-specific prompting protocols that guide the
model through systematic geometric interpretation:
\begin{itemize}
    \item \textbf{Continuous plots (line/scatter):} Prompts require axis-scale verification, coordinate anchoring, localization of explicit markers rather than interpolated lines, and separation of overlapping series using marker shapes or line styles.
    \item \textbf{Categorical plots (bar):} Prompts anchor value estimation at the geometric center of each bar endpoint, using the corresponding horizontal or vertical baseline.
\end{itemize}

Each chart image, together with its structured extraction
instructions (as detailed in Appendix~\ref{app:extraction_instructions}), is submitted to a vision-language model
at native resolution. The model returns the reconstructed
measurements in a standardized Markdown table:
\begin{equation*}
    \text{\small \texttt{| Series Name | X Value / Unit | Y Value / Unit |}}
\end{equation*}
A rule-based parser extracts tabular outputs beginning at the initial delimiter (\texttt{|}) and discards incidental chat tokens. Failed transactions or malformed outputs are recorded with their task parameters to support pipeline resumption. Reconstructed measurements are then associated with caption-derived metadata, including experimental variables, units, assay methods, sample information, etc. (as detailed in Appendix~\ref{app:metadata_prompt}).

\subsection{Batch Execution}
\label{subsec:batch_execution}

To transition from ad-hoc figure parsing to literature-scale corpus harvesting, the workflow is engineered as a batch-processing system:
\begin{itemize}
    \item \textbf{Directory-Level Batch Ingestion:} The system traverses nested corpus directories, autonomously resolving multi-page layouts and establishing hierarchical output structures.
    \item \textbf{Cascaded Candidate Filtering:} Literature harvesting inherently involves high class imbalance (only a fraction of figures report quantitative target assays). The combination of the first-stage co-occurrence filter and Swin Transformer gating serves as a cascaded filtering mechanism, filtering out non-chart panels and irrelevant assays before dispatching expensive multimodal API calls.
    \item \textbf{Stateful Failure Logging and Automated Recovery:} The execution engine dynamically logs transaction anomalies into a structured alongside error traces, supporting systematic recovery across large PDF corpora while avoiding redundant computation.
    \item \textbf{Standardized Dataset Serialization for AI4Science:} Successful extractions are aggregated alongside document provenance and caption schemas into machine-readable Markdown tables and JSON records, directly formatted as instruction-tuning pairs or tabular pre-training corpora.
\end{itemize}

\section{Illustrative Case Studies}
\label{sec:illustrative_examples}

We illustrate the practical execution of the proposed workflow using two biomedical case studies: a high-throughput nanomaterial cytotoxicity screen and a lipid nanoparticle (LNP) delivery study. These examples demonstrate the end-to-end pipeline from document-level figure detection to multimodal reconstruction across heterogeneous chart formats and experimental schemas.

\subsection{Batch Processing Scalability and Funnel Statistics}
\label{subsec:batch_funnel}

To demonstrate batch-processing feasibility, we deployed
\textit{SciFig2Data} on two pilot literature corpora:
605 toxicology and pharmacology articles and 551 LNP-related
articles. As summarized in Table~\ref{tab:funnel_stats},
the cascaded pipeline progressively filters irrelevant
graphical elements and reduces the number of candidates
requiring computationally expensive multimodal reconstruction.

\begin{table}[t]
\centering
\caption{Batch-processing statistics across the toxicology and LNP literature corpora.}
\label{tab:funnel_stats}
\begin{tabular}{lcc}
\toprule
\textbf{Pipeline Stage} & \textbf{Toxicology} & \textbf{LNP} \\
\midrule
Input PDFs (articles)              & 605 & 551 \\
Detected Figures                   & 4630 & 3430 \\
Semantic OCR Filter Passed         & 236 & 224 \\
Decomposed Sub-panels              & 777 & 2213 \\
Classified and Filtered Charts     & 473 & 242 \\
Successful MLLM Extractions        & 468 & 242 \\
\bottomrule
\end{tabular}
\end{table}

\subsection{Case Study 1: Nanomaterial Cytotoxicity Screening}
\label{subsec:cytotoxicity_case_study}

\begin{figure*}[t]
\centering
\resizebox{0.85\textwidth}{!}{%
\begin{tikzpicture}[
    font=\small\sffamily,
    box/.style={draw=gray!60, fill=white, rounded corners=4pt, line width=0.7pt, inner sep=4pt, align=center},
    header/.style={fill=blue!8, draw=blue!40, rounded corners=3pt, font=\bfseries\footnotesize, inner sep=3pt, text=blue!80!black},
    arr/.style={-{Stealth[length=2.2mm,width=1.5mm]}, line width=1.1pt, draw=blue!75!black},
    subarr/.style={-{Stealth[length=2mm,width=1.3mm]}, line width=0.9pt, dashed, draw=orange!80!black},
    tablebox/.style={fill=gray!3, draw=gray!40, rounded corners=4pt, inner sep=4pt}
]

\node[header] (h1) at (0, 0) {Document Layout Analysis};
\node[box, below=0.15cm of h1] (n1) {
    \includegraphics[height=4.6cm]{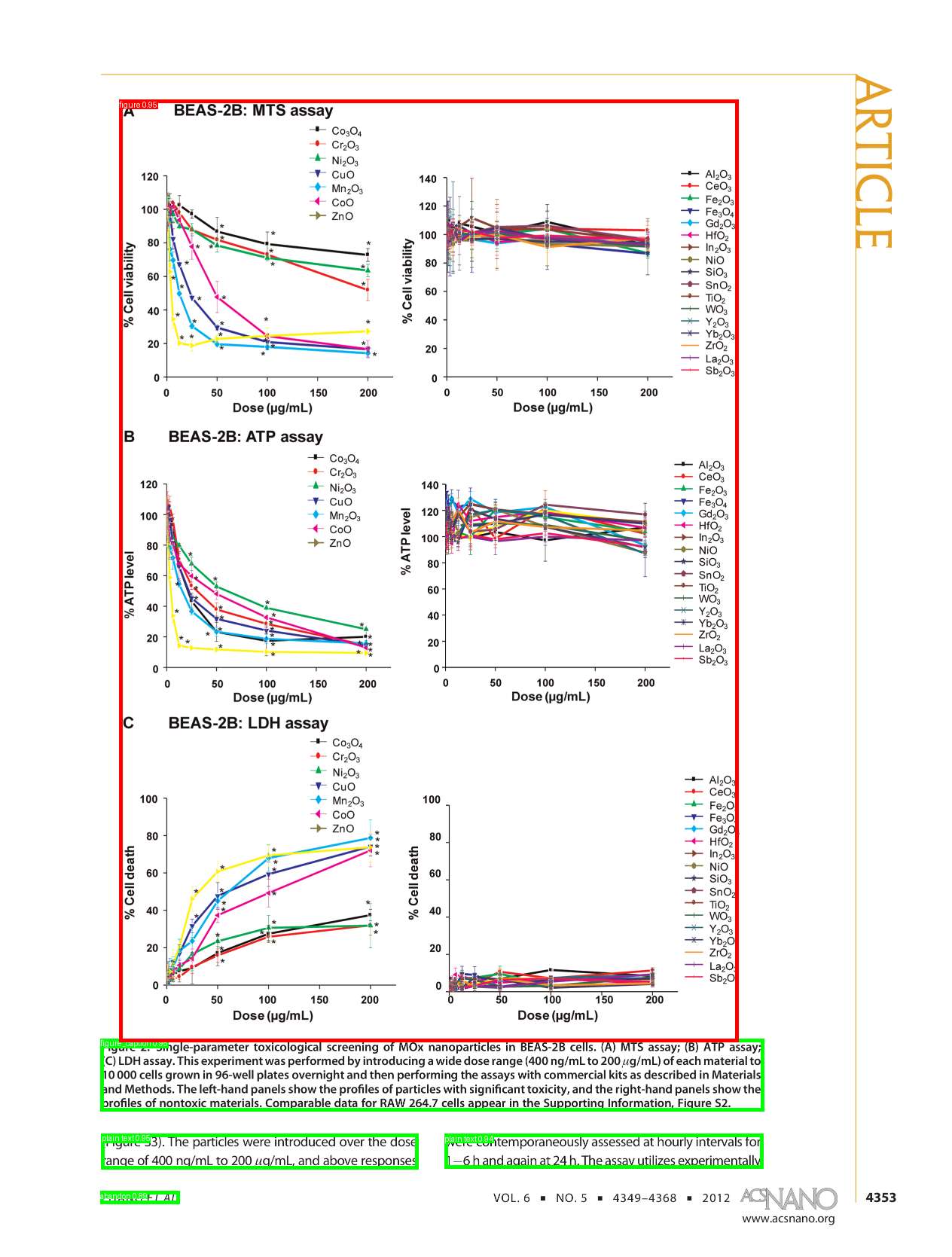}
};
\node[below=0.06cm of n1, font=\scriptsize\color{black!70}] (lbl1) {};

\node[header] (h3) at (5.3, 0) {Compound Chart Separation};
\node[box, below=0.15cm of h3] (n3) {
    \includegraphics[height=4.6cm]{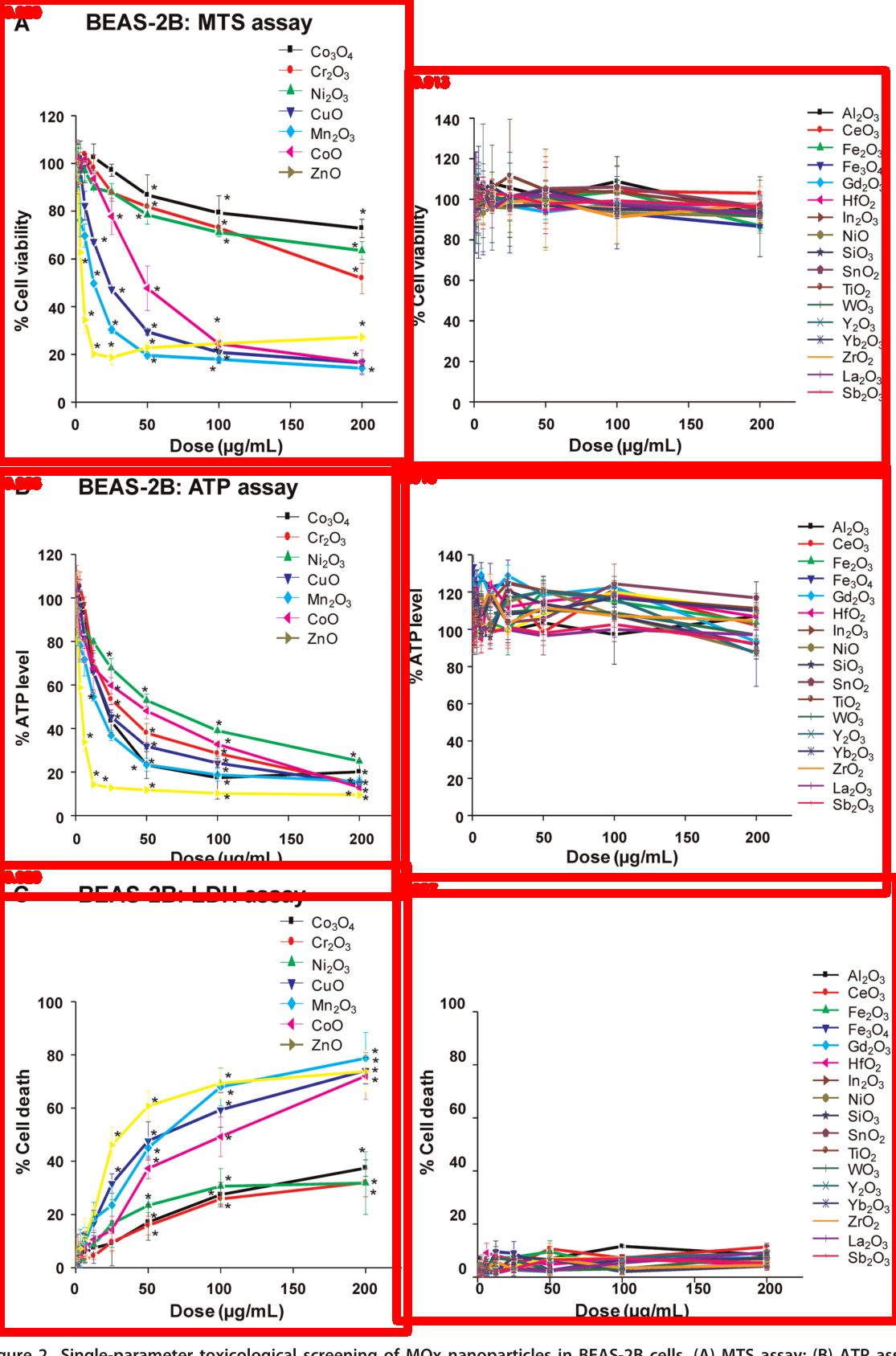}
};
\node[below=0.06cm of n3, font=\scriptsize\color{black!70}] {};

\node[header] (h4) at (10.4, 0) {Chart Classification};
\node[box, below=0.15cm of h4, text width=3.8cm] (n4) {
    \includegraphics[height=3.1cm]{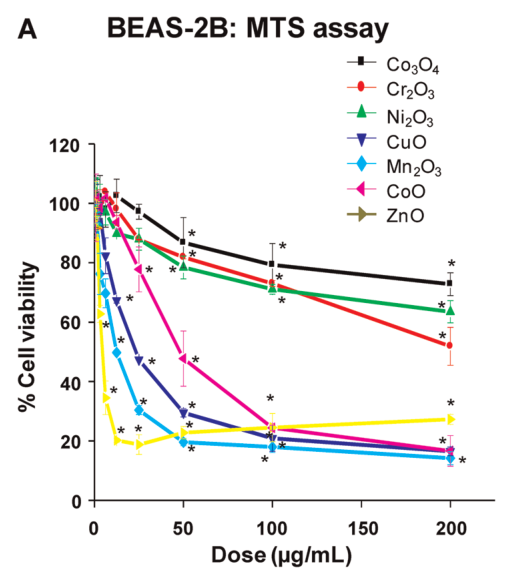}\\[3pt]
    \textbf{\footnotesize Sub-panel A (BEAS-2B: MTS)}\\[1pt]
    \color{blue!80!black}\textbf{\scriptsize Class: Line Chart}
};

\node[header] (h2) at (0, -6.4) {Caption OCR \& Semantic Filter};
\node[box, below=0.15cm of h2, text width=4.6cm] (n2) {
    \includegraphics[width=4.4cm]{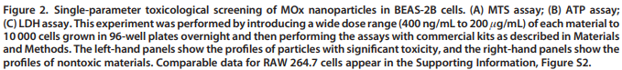}\\[3pt]
    \rule{4.4cm}{0.4pt}\\[2pt]
    \centering
    \textbf{\footnotesize Lexical Screening Passed:}\\[1pt]
    \scriptsize $\mathcal{K}_{\text{indep}}: \text{``dose''} \quad \mathcal{K}_{\text{dep}}: \text{``viability''}$
};

\node[header] (h5) at (8.2, -5.8) {Multi-modal Table Reconstruction};
\node[tablebox, below=0.15cm of h5, text width=5.8cm] (n5) {
    \textbf{\scriptsize (A) Instruction-Guided Visual Data Extraction:}
    \vspace{1pt}

    \tiny
    \setlength{\tabcolsep}{2pt}
    \begin{tabular*}{\linewidth}{@{\extracolsep{\fill}}lccc@{}}
    \toprule
    \textbf{Series} & \textbf{Dose ($\mu\text{g/mL}$)} & \textbf{Viability (\%)} \\
    \midrule
    $\text{Co}_3\text{O}_4$ & 200 & 73.5 \\
    $\text{ZnO}$ & 12.5 & 20.3 \\
    $\text{...}$ & ... & ... \\
    \bottomrule
    \end{tabular*}

    \vspace{2pt}
    \textbf{\scriptsize (B) Caption-Enriched Schema Integration:}
    \vspace{1pt}

    \tiny
    \setlength{\tabcolsep}{2pt}
    \begin{tabular*}{\linewidth}{@{\extracolsep{\fill}}lcccc@{}}
    \toprule
    \textbf{Nanoparticle} & \textbf{Dose} & \textbf{Viability} & \textbf{Assay} & \textbf{Cell} \\
    \midrule
    $\text{Co}_3\text{O}_4$ & $200\,\mu\text{g/mL}$ & $73.5\%$ & MTS & BEAS-2B \\
    $\text{ZnO}$ & $12.5\,\mu\text{g/mL}$ & $20.3\%$ & MTS & BEAS-2B \\
    $\text{...}$ & $\text{...}$ & $...$ & ... & ... \\
    \bottomrule
    \end{tabular*}
};

\draw[arr] (lbl1.south) -- (h2.north) 
    node[midway, right=4pt, font=\scriptsize] {Image \& Caption};
\draw[arr] (n1.east) -- (n3.west) 
    node[midway, above=2pt, font=\scriptsize] {Passed};
\draw[arr] (n3.east |- n3.center) -- (n4.west |- n3.center) 
    node[midway, above=2pt, font=\scriptsize] {Sub-panels};
\draw[arr] (n4.south) -- (n4.south |- n5.north) 
    node[midway, right=4pt, font=\scriptsize, align=left] {Visual\\Reasoning};
\draw[subarr] (n2.east) -- ++(1.3, 0) |- (n5.west) 
    node[pos=0.25, above=1pt, font=\scriptsize\color{orange!90!black}] {Metadata Contextual Enrichment};

\end{tikzpicture}
}
\caption{Workflow execution on a compound figure from \textit{ACS Nano}. The pipeline demonstrates layout detection, caption filtering, panel segmentation, chart classification, and multimodal data extraction with context enrichment.}
\label{fig:case_study_walkthrough}
\end{figure*}

\paragraph{Source Figure and Workflow Execution}
The first study, published in \textit{ACS Nano}~\cite{toxicology_case_study}, assesses the cytotoxicity of 24 metal oxide ($\text{MO}_x$) nanoparticles in human bronchial epithelial cells (BEAS-2B). As shown in Fig.~\ref{fig:case_study_walkthrough}, Figure~2 of the source document is a compound figure comprising six panels spanning three biological endpoints (MTS cell viability, ATP activity, and LDH release). 

The layout detector (YOLOv10) located the figure and its caption with high confidence ($\tau_{\text{conf}} \ge 0.8$). Lexical screening confirmed target keywords (e.g., ``dose'', ``concentration'', ``viability''). Following YOLOv2-based subfigure separation and subsequent boundary refinement, sub-panel A (BEAS-2B: MTS assay) was classified as a \texttt{line} chart ($p = 0.96$), activating the corresponding coordinate-based extraction pipeline.

\paragraph{Data Reconstruction and Contextual Integration}
The multimodal extraction engine (\texttt{gemini-3.1-pro}~\cite{google2026gemini31}) resolved the series markers and axes, recovering representative measurements such as $20.3 \pm 3.1\%$ for $\text{ZnO}$ at $12.5\,\mu\text{g/mL}$ and $73.5 \pm 5.2\%$ for $\text{Co}_3\text{O}_4$ at $200\,\mu\text{g/mL}$. Caption parsing linked these measurements directly to experimental context (MTS assay, BEAS-2B cells), transforming raw chart visual tokens into structured toxicological records.

\subsection{Case Study 2: LNP-Mediated mRNA Delivery}
\label{subsec:lnp_case_study}

\begin{figure*}[t]
\centering
\resizebox{0.85\textwidth}{!}{%
\begin{tikzpicture}[
    font=\small\sffamily,
    box/.style={draw=gray!60, fill=white, rounded corners=4pt, line width=0.7pt, inner sep=4pt, align=center},
    header/.style={fill=blue!8, draw=blue!40, rounded corners=3pt, font=\bfseries\footnotesize, inner sep=3pt, text=blue!80!black},
    arr/.style={-{Stealth[length=2.2mm,width=1.5mm]}, line width=1.1pt, draw=blue!75!black},
    subarr/.style={-{Stealth[length=2mm,width=1.3mm]}, line width=0.9pt, dashed, draw=orange!80!black},
    tablebox/.style={fill=gray!3, draw=gray!40, rounded corners=4pt, inner sep=4pt}
]

\node[header] (h1) at (0, 0) {Document Layout Analysis};
\node[box, below=0.15cm of h1] (n1) {
    \includegraphics[height=4.6cm]{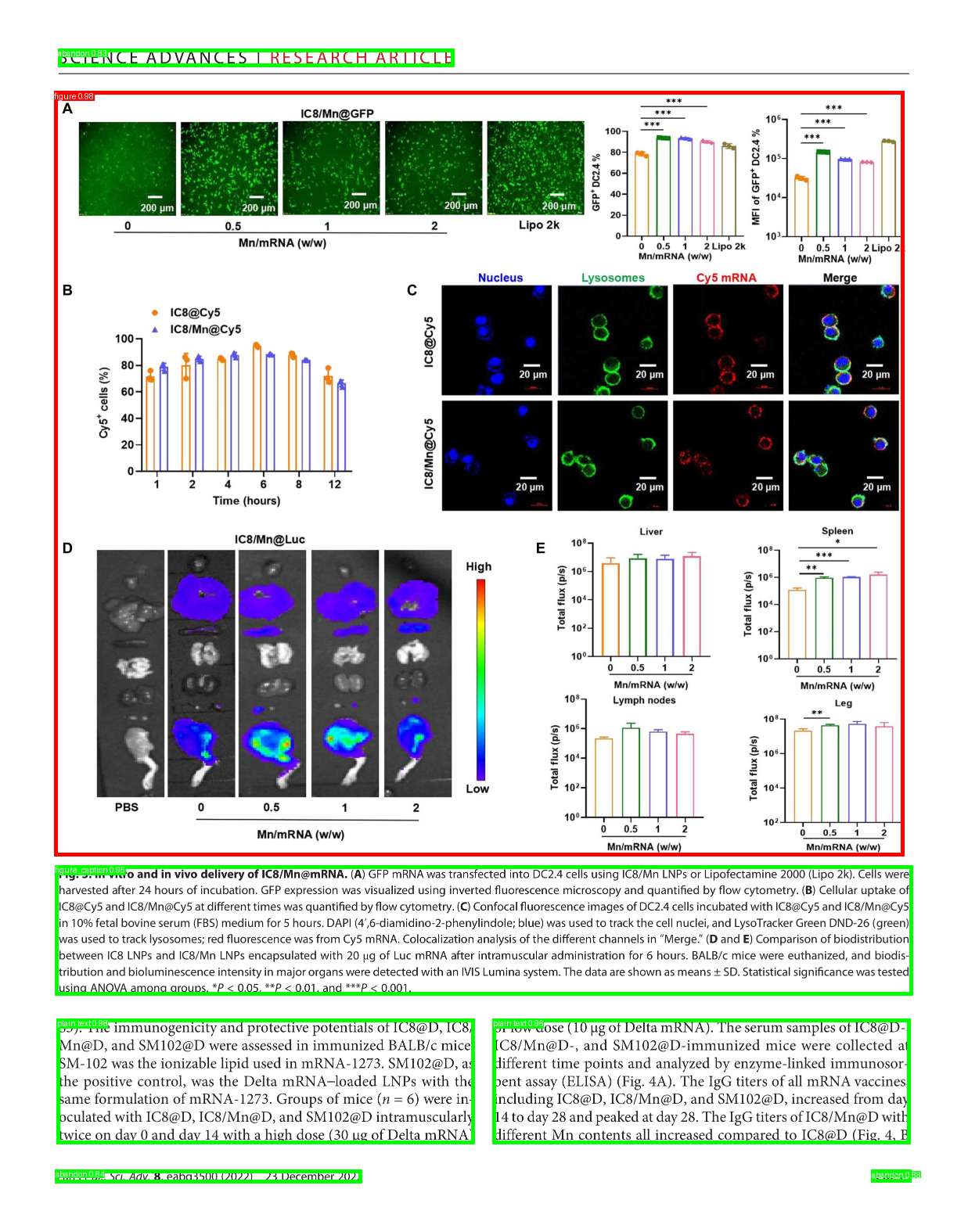}
};
\node[below=0.06cm of n1, font=\scriptsize\color{black!70}] (lbl1) {};

\node[header] (h3) at (5.8, 0) {Compound Chart Separation};
\node[box, below=0.15cm of h3] (n3) {
    \includegraphics[height=4.6cm]{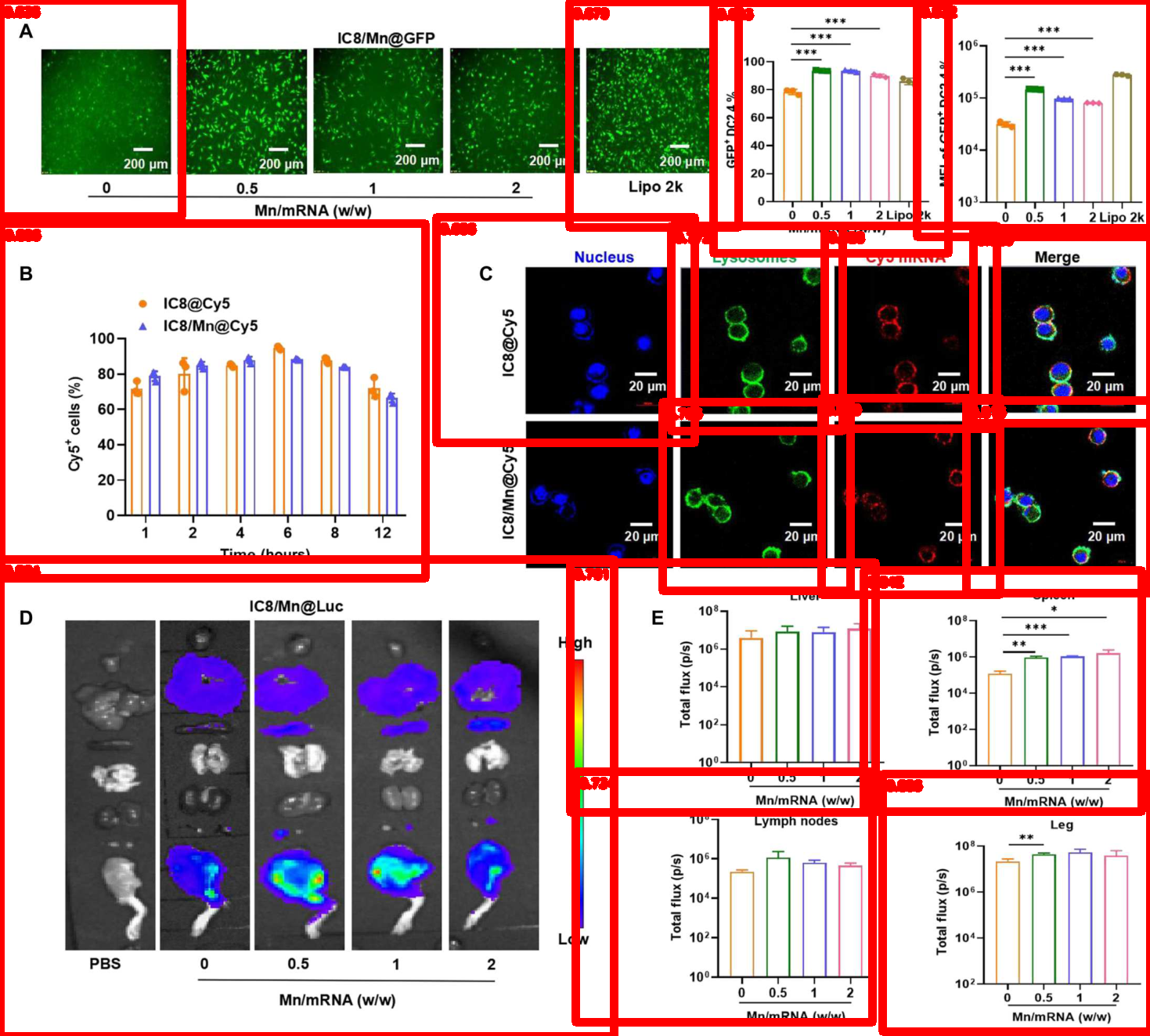}
};
\node[below=0.06cm of n3, font=\scriptsize\color{black!70}] {};

\node[header] (h4) at (12.3, 0) {Chart Classification};
\node[box, below=0.15cm of h4, text width=3.8cm] (n4) {
    \includegraphics[height=3.1cm]{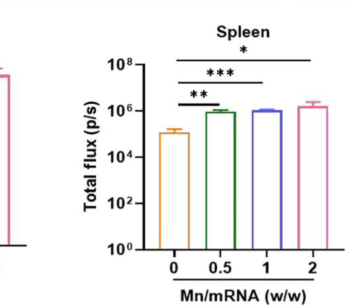}\\[3pt]
    \textbf{\footnotesize Sub-panel E (Spleen: Biodistribution)}\\[1pt]
    \color{blue!80!black}\textbf{\scriptsize Class: Vertical-bar Chart}
};

\node[header] (h2) at (0, -6.9) {Caption OCR \& Semantic Filter};
\node[box, below=0.15cm of h2, text width=4.6cm] (n2) {
    \includegraphics[width=4.4cm]{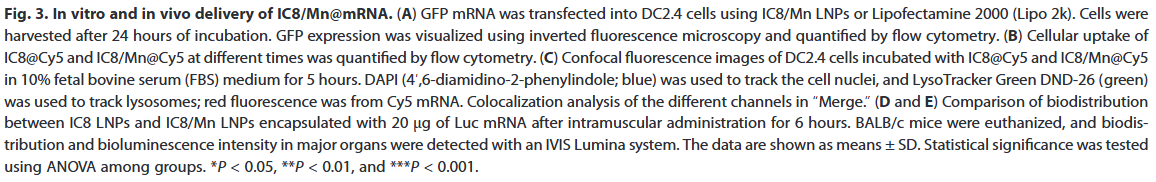}\\[3pt]
    \rule{4.4cm}{0.4pt}\\[2pt]
    \centering
    \textbf{\footnotesize Lexical Screening Passed:}\\[1pt]
    \scriptsize $\mathcal{K}_{\text{indep}}: \text{``mRNA''} \quad \mathcal{K}_{\text{dep}}: \text{``flux''}$
};

\node[header] (h5) at (8.2, -6.3) {Multi-modal Table Reconstruction};
\node[tablebox, anchor=north west, text width=8.9cm] (n5) at ([xshift=-3.04cm, yshift=-0.15cm]h5.south) {
    \textbf{\scriptsize (A) Instruction-Guided Visual Data Extraction:}
    \vspace{1pt}

    \tiny
    \setlength{\tabcolsep}{1.5pt}
    \begin{tabular*}{\linewidth}{@{\extracolsep{\fill}}lcc@{}}
    \toprule
    \textbf{Series Name} & \textbf{X Value / Unit (w/w)} & \textbf{Y Value / Unit (p/s)}\\
    \midrule
    IC8/Mn LNP & 0 & 5.0e+04\\
    IC8/Mn LNP & 0.5 & 8.0e+05\\
    IC8/Mn LNP & 1 & 1.0e+06\\
    IC8/Mn LNP & 2 & 1.8e+06\\
    \bottomrule
    \end{tabular*}

    \vspace{2pt}
    \textbf{\scriptsize (B) Caption-Enriched Schema Integration:}
    \vspace{1pt}

    \tiny
    \setlength{\tabcolsep}{1.5pt}
    \begin{tabular*}{\linewidth}{@{\extracolsep{\fill}}lcccccc@{}}
    \toprule
    \textbf{Organ} &
    \begin{tabular}{@{}c@{}}\textbf{Mn/mRNA}\\\textbf{Ratio (w/w)}\end{tabular} &
    \textbf{LNP Type} &
    \textbf{mRNA Type} &
    \begin{tabular}{@{}c@{}}\textbf{mRNA}\\\textbf{Amount (µg)}\end{tabular} &
    \begin{tabular}{@{}c@{}}\textbf{Administration}\\\textbf{Time (h)}\end{tabular} &
    \textbf{Flux (p/s)}\\
    \midrule
    Spleen & 0   & IC8 LNP    & Luc & 20 & 6 & 5.0E+04\\
    Spleen & 0.5 & IC8/Mn LNP & Luc & 20 & 6 & 8.0E+05\\
    Spleen & 1   & IC8/Mn LNP & Luc & 20 & 6 & 1.0E+06\\
    Spleen & 2   & IC8/Mn LNP & Luc & 20 & 6 & 1.8E+06\\
    \bottomrule
    \end{tabular*}
};

\draw[arr] (lbl1.south) -- (h2.north) node[midway, right=4pt, font=\scriptsize] {Image \& Caption};
\draw[arr] (n1.east) -- (n3.west) node[midway, above=2pt, font=\scriptsize] {Passed};
\draw[arr] (n3.east |- n3.center) -- (n4.west |- n3.center) node[midway, above=2pt, font=\scriptsize] {Sub-panels};
\draw[arr] (n4.south) -- (n4.south |- n5.north) node[midway, right=4pt, font=\scriptsize, align=left] {Visual\\Reasoning};
\draw[subarr] (n2.east) -- ++(1.3, 0) |- (n5.west) node[pos=0.25, above=1pt, font=\scriptsize\color{orange!90!black}] {Metadata Contextual Enrichment};

\end{tikzpicture}
}
\caption{Workflow execution on an LNP-mediated mRNA delivery study from \textit{Science Advances}. The pipeline handles heterogeneous subfigure formats, identifies a bar chart (Panel E, Spleen), and supports numerical reconstruction from bar charts enriched with formulation, organ, and dosage metadata.}
\label{fig:lnp_case_study}
\end{figure*}

\paragraph{Heterogeneous Panel Handling and Filtering}
To demonstrate the workflow's applicability to heterogeneous figure structures, we evaluated an in vivo mRNA delivery study from \textit{Science Advances}~\cite{lnp_case_study}. Unlike the previous homogeneous screen, Figure~3 in this study contains five multi-modal panels (A--E) encompassing confocal microscopy, flow cytometry, bioluminescence imaging, and quantitative vertical-bar charts (Fig.~\ref{fig:lnp_case_study}). Document layout analysis identified the figure and caption, while OCR filtering detected delivery-specific cues (e.g., ``bioluminescence'', ``Mn/mRNA ratio''). Panel segmentation isolated individual subfigures, and chart classification successfully distinguished quantitative bar plots from adjacent non-chart biological images.

\paragraph{Data Reconstruction and Contextual Integration}
For numerical reconstruction, sub-panel E (spleen biodistribution) was selected as a representative example. This panel reports total bioluminescent flux across four Mn/mRNA weight ratios ($0$, $0.5$, $1$, and $2$), with the vertical axis displayed on a logarithmic scale. The extraction module identified the axis scaling and interpreted the bar heights accordingly, enabling the reconstruction of the corresponding flux values without assuming a linear relationship between pixel positions and numerical values.

As illustrated in Fig.~\ref{fig:lnp_case_study}, the reconstructed measurements were first organized into a chart-level table containing series names, Mn/mRNA ratios, and flux values. Caption-derived experimental metadata were then integrated to generate context-enriched records, associating each measurement with its target organ (spleen), LNP formulation, mRNA type (Luc), administered mRNA amount (20~$\mu$g), and administration time (6~h). This integration transforms isolated chart measurements into structured records that retain key experimental context and are more readily usable for downstream analysis.

Together, these case studies illustrate the applicability
of the proposed pipeline across different biological
domains, chart typologies (line and bar charts), and compound figures combining
imaging and quantitative data.

\section{Discussion}
\label{sec:discussion}

\textit{SciFig2Data} addresses scientific figure recovery as an integrated document-to-data task rather than an isolated chart digitization problem. The two case studies illustrate complementary challenges in scholarly documents: the nanomaterial cytotoxicity study contains multiple related quantitative panels, whereas the LNP delivery study combines imaging modalities with quantitative plots, including a bar chart. Across these examples, the workflow connects document-level figure localization and panel organization with chart identification, numerical reconstruction, and caption-based recovery of experimental context. This integration is important because extracted values are difficult to reuse, interpret, or compare when separated from their experimental conditions.

Beyond individual figure reconstruction, \textit{SciFig2Data} is designed to support the construction of structured scientific datasets for AI for Science (AI4Science). Applications such as molecular property prediction, quantitative structure--activity relationship (QSAR) modeling, and toxicological risk assessment can benefit from larger and more accessible experimental data resources. By transforming compound PDF figures into structured records that associate figure content, captions, and reconstructed numerical measurements, the workflow provides a potential source of training and reference data for scientific machine learning. Such records could also support the development of scientific vision-language models and domain-specific predictive models, although their suitability for these applications requires further validation of extraction accuracy, metadata consistency, and data quality.

The workflow's preprocessing stages are intended to reduce visual and semantic interference before numerical reconstruction. Multi-scale rendering and boundary expansion help retain small annotations, while panel decomposition and reading-order restoration provide focused inputs for downstream analysis. Semantic OCR filtering and chart classification further narrow the candidate panels, and chart-specific instructions guide the reconstruction of different plot types. The LNP case also highlights the importance of recognizing logarithmic axes when converting visual positions into numerical values. However, errors in detection, segmentation, filtering, or chart interpretation can propagate through subsequent stages.

Several limitations remain. First, although the workflow has been demonstrated on two biomedical literature corpora, systematic quantitative evaluation of extraction accuracy across diverse scientific publications remains necessary to establish its generalizability and numerical fidelity. Second, the current implementation covers five chart paradigms (horizontal bar, vertical bar, line, scatter, and scatter-line), while other scientific visualizations remain outside its scope. Third, semantic filtering depends on domain-specific lexicons and may miss figures using uncommon terminology, abbreviations, or alternative experimental descriptions. Finally, the extracted tables still require human-in-the-loop post-processing, particularly for synonym normalization and the consolidation of related entries, before they can be used in downstream analyses. Future improvements should therefore focus on broader evaluation, expanded chart coverage, cross-domain semantic filtering, and source-level provenance recording to improve the traceability and validation of extracted measurements.

\subsection*{Reproducibility Statement}

The source code of \textit{SciFig2Data} will be made publicly available at \url{https://github.com/ISSAIR/SciFig2Data} to facilitate reproducibility and further research.

\bibliographystyle{apalike}
\bibliography{references}

\begin{thebibliography}{}

\bibitem[Davila et~al., 2021]{davila2021chart}
Davila, K., Setlur, S., Doermann, D., Kota, B.~U., and Govindaraju, V. (2021).
\newblock Chart mining: A survey of methods for automated chart analysis.
\newblock {\em IEEE Transactions on Pattern Analysis and Machine Intelligence}, 43(11):3799--3819.

\bibitem[Dhote et~al., 2024]{dhote2024swinchart}
Dhote, A., Javed, M., and Doermann, D.~S. (2024).
\newblock Swin-chart: An efficient approach for chart classification.
\newblock {\em Pattern Recognition Letters}, 185:203--209.

\bibitem[Fan et~al., 2022]{lnp_case_study}
Fan, N., Chen, K., Zhu, R., Zhang, Z., Huang, H., Qin, S., Zheng, Q., He, Z., He, X., Xiao, W., Zhang, Y., Gu, Y., Zhao, C., Liu, Y., Jiang, X., Li, S., Wei, Y., and Song, X. (2022).
\newblock Manganese-coordinated mrna vaccines with enhanced mrna expression and immunogenicity induce robust immune responses against sars-cov-2 variants.
\newblock {\em Science Advances}, 8(51):eabq3500.

\bibitem[{Google DeepMind}, 2026]{google2026gemini31}
{Google DeepMind} (2026).
\newblock Gemini 3.1 pro model card.
\newblock \url{https://deepmind.google/models/model-cards/gemini-3-1-pro/}.

\bibitem[Islam et~al., 2024]{islam2024chartcomprehension}
Islam, M.~S., Rahman, R., Masry, A., Laskar, M. T.~R., Nayeem, M.~T., and Hoque, E. (2024).
\newblock Are large vision language models up to the challenge of chart comprehension and reasoning?
\newblock In {\em Findings of the Association for Computational Linguistics: EMNLP 2024}, pages 3334--3368.

\bibitem[Liu et~al., 2023]{liu2023deplot}
Liu, F., Eisenschlos, J., Piccinno, F., Krichene, S., Pang, C., Lee, K., Joshi, M., Chen, W., Collier, N., and Altun, Y. (2023).
\newblock Deplot: One-shot visual language reasoning by plot-to-table translation.
\newblock In {\em Findings of the Association for Computational Linguistics: ACL 2023}, pages 10381--10399.

\bibitem[Liu et~al., 2026]{liu2026start}
Liu, Z., Gao, X., Niu, F., Gao, Q., Liu, L., and Piramuthu, R. (2026).
\newblock Start: Spatial and textual learning for chart understanding.
\newblock In {\em 2026 IEEE/CVF Winter Conference on Applications of Computer Vision (WACV)}, pages 8146--8156. IEEE.

\bibitem[Liu et~al., 2021]{liu2021swin}
Liu, Z., Lin, Y., Cao, Y., Hu, H., Wei, Y., Zhang, Z., Lin, S., and Guo, B. (2021).
\newblock Swin transformer: Hierarchical vision transformer using shifted windows.
\newblock In {\em Proceedings of the IEEE/CVF International Conference on Computer Vision (ICCV)}, pages 10012--10022.

\bibitem[Luo et~al., 2021]{luo2021chartocr}
Luo, J., Li, Z., Wang, J., and Lin, C.-Y. (2021).
\newblock Chartocr: Data extraction from charts images via a deep hybrid framework.
\newblock In {\em Proceedings of the IEEE/CVF Winter Conference on Applications of Computer Vision}, pages 1917--1925.

\bibitem[Pfitzmann et~al., 2022]{pfitzmann2022doclaynet}
Pfitzmann, B., Auer, C., Dolfi, M., Nassar, A.~S., and Staar, P. W.~J. (2022).
\newblock Doclaynet: A large human-annotated dataset for document-layout analysis.

\bibitem[Pick et~al., 2019]{pick2019metadigitise}
Pick, J.~L., Nakagawa, S., and Noble, D. W.~A. (2019).
\newblock Reproducible, flexible and high-throughput data extraction from primary literature: The metadigitise r package.
\newblock {\em Methods in Ecology and Evolution}, 10(3):426--431.

\bibitem[Redmon and Farhadi, 2017]{redmon2017yolo9000}
Redmon, J. and Farhadi, A. (2017).
\newblock Yolo9000: Better, faster, stronger.
\newblock In {\em Proceedings of the IEEE Conference on Computer Vision and Pattern Recognition (CVPR)}, pages 7263--7271.

\bibitem[Tsutsui and Crandall, 2017]{tsutsui2017compound}
Tsutsui, S. and Crandall, D. (2017).
\newblock A data driven approach for compound figure separation using convolutional neural networks.
\newblock In {\em 2017 14th IAPR International Conference on Document Analysis and Recognition (ICDAR)}.

\bibitem[Wang et~al., 2023]{wang2023scientific}
Wang, H., Fu, T., Du, Y., Gao, W., Huang, K., Liu, Z., Chandak, P., Liu, S., Van~Katwyk, P., Deac, A., Anandkumar, A., Bergen, K., Gomes, C.~P., Ho, S., Kohli, P., Lasenby, J., Leskovec, J., Liu, T.-Y., Manrai, A., Marks, D., Ramsundar, B., Song, L., Sun, J., Tang, J., Veli{\v{c}}kovi{\'{c}}, P., Welling, M., Zhang, L., Coley, C.~W., Bengio, Y., and Zitnik, M. (2023).
\newblock Scientific discovery in the age of artificial intelligence.
\newblock {\em Nature}, 620(7972):47--60.

\bibitem[Wu and Jin, 2026]{wu2026chartdensity}
Wu, X. and Jin, Y. (2026).
\newblock Chartdensity-bench: Benchmarking mllms for numerical data reconstruction under visual density.

\bibitem[Yuan et~al., 2026]{yuan2026chartrecover}
Yuan, Y., Liang, S., Zhang, J., Wang, Y., and Wang, Z. (2026).
\newblock A deep learning framework for scientific chart data extraction and reconstruction.
\newblock {\em Communications Engineering}, 5(1):147.

\bibitem[Zhang et~al., 2012]{toxicology_case_study}
Zhang, H., Ji, Z., Xia, T., Meng, H., Low-Kam, C., Liu, R., Pokhrel, S., Lin, S., Wang, X., Liao, Y.-P., Wang, M., Li, L., Rallo, R., Damoiseaux, R., Telesca, D., M{\"a}dler, L., Cohen, Y., Zink, J.~I., and Nel, A.~E. (2012).
\newblock Use of metal oxide nanoparticle band gap to develop a predictive paradigm for oxidative stress and acute pulmonary inflammation.
\newblock {\em ACS Nano}, 6(5):4349--4368.

\bibitem[Zhao et~al., 2024]{zhao2024doclayoutyolo}
Zhao, Z., Kang, H., Wang, B., and He, C. (2024).
\newblock Doclayout-yolo: Enhancing document layout analysis through diverse synthetic data and global-to-local adaptive perception.

\end{thebibliography}

\appendix
\section{Chart Data Extraction Instructions}
\label{app:extraction_instructions}

To guide the vision-language model in systematically interpreting heterogeneous chart paradigms during the numerical reconstruction stage, we employ the following precise extraction prompt specifications:

\subsection{Line Chart Data Extraction Instructions}
\begin{itemize}
    \item \textbf{Basic Identification}:
    \begin{itemize}
        \item Axis Analysis: Horizontal/vertical axis labels, units, scale range, and intervals (for unit conversion reference only).
    \end{itemize}
    \item \textbf{Data Point Extraction Rules}:
    \begin{itemize}
        \item \textbf{Key Point Location}: Prioritize using \textbf{actual nodes} on the line as data points. When no nodes exist, use \textbf{uniform sampling} (evenly spaced points based on scale intervals).
        \item \textbf{Baseline Determination}: Use the x-value (horizontal axis value) of data points as the baseline.
    \end{itemize}
    \item \textbf{Specific Scenario Handling Rules}:
    \begin{itemize}
        \item \textbf{Multi-line Processing}: Distinguish data sequences of different lines.
    \end{itemize}
    \item \textbf{Output Format Specification}:
    \begin{quote}
        \texttt{| Series Name | X Value / Unit | Y Value / Unit |}
    \end{quote}
    \textbf{Format Requirements}:
    \begin{itemize}
        \item X Value: Write numerical values if the horizontal axis is numerical, or category names if categorical.
        \item Units are only marked in column headers within parentheses, not repeated in individual cells.
    \end{itemize}
\end{itemize}

\subsection{Bar Chart Data Extraction Instructions}
\begin{itemize}
    \item \textbf{Basic Identification}:
    \begin{itemize}
        \item Chart Type Identification: Clearly distinguish between regular bar charts or stacked bar charts.
        \item Axis Analysis: Horizontal/vertical axis labels, units, scale range, and intervals (for unit conversion reference only).
    \end{itemize}
    \item \textbf{Data Point Extraction Rules}:
    \begin{itemize}
        \item \textbf{Key Point Location}: Use the \textbf{center of each bar's top} as the data point.
        \item \textbf{Baseline Determination}: Use the x-value (horizontal axis value) corresponding to each bar as the baseline.
    \end{itemize}
    \item \textbf{Specific Scenario Handling Rules}:
    \begin{itemize}
        \item \textbf{Stacked Bar Processing}: Record both individual segment values (stacked portions) and total bar height.
        \item \textbf{Error Bar Processing}: Extract error values.
        \item \textbf{Horizontal Bar Charts}: Use y-axis values as the baseline for data point location and X value extraction.
    \end{itemize}
    \item \textbf{Output Format Specification}:
    \begin{quote}
        \texttt{| Series Name | X Value / Unit | Y Value / Unit |}
    \end{quote}
    \textbf{Format Requirements}:
    \begin{itemize}
        \item X Value: Write numerical values if the horizontal axis is numerical, or category names if categorical.
        \item Units are only marked in column headers within parentheses, not repeated in individual cells.
    \end{itemize}
\end{itemize}

\subsection{Scatter Plot Data Extraction Instructions}
\begin{itemize}
    \item \textbf{Basic Identification}:
    \begin{itemize}
        \item Axis Analysis: Horizontal/vertical axis labels, units, scale range, and intervals (for unit conversion reference only).
    \end{itemize}
    \item \textbf{Data Point Extraction Rules}:
    \begin{itemize}
        \item \textbf{Key Point Location}: Extract all visible scatter points as data points. For dense regions, prioritize characteristic distribution points (such as edge points, cluster center points).
        \item \textbf{Baseline Determination}: Use the x-value (horizontal axis value) of data points as the baseline.
    \end{itemize}
    \item \textbf{Specific Scenario Handling Rules}:
    \begin{itemize}
        \item \textbf{Multi-series Processing}: Distinguish data sequences by different colors/shapes.
        \item \textbf{Overlapping Point Processing}: Identify and annotate overlapping or near-overlapping data points.
    \end{itemize}
    \item \textbf{Output Format Specification}:
    \begin{quote}
        \texttt{| Series Name | X Value / Unit | Y Value / Unit |}
    \end{quote}
    \textbf{Format Requirements}:
    \begin{itemize}
        \item X Value: Write numerical values if the horizontal axis is numerical, or category names if categorical.
        \item Units are only marked in column headers within parentheses, not repeated in individual cells.
    \end{itemize}
\end{itemize}

\section{Caption-Based Metadata Integration Instructions}
\label{app:metadata_prompt}

To guide the multimodal large language model in fusing visual chart evidence with textual caption metadata during the final reconstruction stage, we employ the following structured prompt instructions:

\begin{itemize}
    \item \textbf{Input Modality}: Multi-modal input comprising the \textbf{target chart image} and the extracted text caption.
    \item \textbf{Core Task Requirements}:
    \begin{itemize}
        \item \textbf{Visual Analysis}: Analyze the provided chart image to identify chart legends, labels, and visual data trends.
        \item \textbf{Caption Metadata Parsing}: Identify specific experimental configurations, parameters, and conditions described in the figure caption.
    \end{itemize}
    \item \textbf{Table Structure Optimization}:
    \begin{itemize}
        \item Accurately name column headers based on chart content and axis units.
        \item Add necessary contextual columns when appropriate (e.g., time, assay type, nanoparticle/material type, cell type, or formulation parameters).
        \item Incorporate correct units extracted from both axis labels and caption text.
    \end{itemize}
    \item \textbf{Output Requirements}:
    \begin{itemize}
        \item Preserve the original document structure and hierarchy.
        \item Modify and enrich only the target table sections.
        \item Output the fully optimized Markdown file containing context-enriched experimental data.
    \end{itemize}
\end{itemize}

\end{document}